\documentclass[letterpaper, 10 pt, conference]{ieeeconf}  

\IEEEoverridecommandlockouts                              
\usepackage{bmathshortcuts}

\usepackage{algorithmic}
\usepackage[ruled,vlined]{algorithm2e}
\usepackage{amsmath,amssymb,amsfonts}
\usepackage{bm}
\usepackage{booktabs}
\usepackage{cite}
\usepackage{color}
\usepackage{comment}
\usepackage{fancyhdr}
\usepackage{float}
\usepackage{gensymb}
\usepackage{graphicx}
\usepackage[hidelinks]{hyperref} 
\usepackage{mathrsfs}
\usepackage{soul}
\usepackage{textcomp}
\usepackage{times}
\usepackage[usenames,dvipsnames]{xcolor} 
\usepackage{url}
\DeclareMathOperator*{\argmax}{argmax}

{\vspace{-\topsep}\begin{itemize}\itemsep1pt \parskip0pt \parsep1pt}
{\end{itemize}\vspace{-\topsep}}

{\vspace{-\topsep}\begin{enumerate}\itemsep1pt \parskip0pt \parsep1pt}
{\end{enumerate}\vspace{-\topsep}}

\begin{document}

\title{\LARGE \bf
Self-Sensing for Proprioception and Contact Detection in Soft Robots Using Shape Memory Alloy Artificial Muscles}



\author{Ran Jing$^{1}$, Meredith L. Anderson$^{1}$, Juan C. Pacheco Garcia$^{1}$, Andrew P. Sabelhaus$^{1,2}$
\thanks{$^1$R. Jing, M.L. Anderson, J.C. Pacheco Garcia, and A.P. Sabelhaus are with the Department of Mechanical Engineering, Boston University, Boston MA, USA. {\tt\small \{rjing, merland, jcp29, asabelha\}@bu.edu}. $^2$A.P. Sabelhaus is also with the Division of Systems Engineering, Boston University, Boston MA, USA. }
}
\maketitle

\pagestyle{empty}  
\thispagestyle{empty} 

\begin{abstract}

Estimating a soft robot's pose and applied forces, also called proprioception, is crucial for safe interaction of the robot with its environment. 
However, most solutions for soft robot proprioception use dedicated sensors, particularly for external forces, which introduce design trade-offs, rigidity, and risk of failure.
This work presents an approach for pose estimation and contact detection for soft robots actuated by shape memory alloy (SMA) artificial muscles, using no dedicated force sensors.
Our framework uses the unique material properties of SMAs to self-sense their internal stress, via offboard measurements of their electrical resistance and in-situ temperature readings, in an existing fully-soft limb design.
We demonstrate that a simple polynomial regression model on these measurements is sufficient to predict the robot's pose, under no-contact conditions.
Then, we show that if an additional measurement of the true pose is available (e.g. from an already-in-place bending sensor), it is possible to predict a binary contact/no-contact using multiple combinations of self-sensing signals.
Our hardware tests verify our hypothesis via a contact detection test with a human operator.
This proof-of-concept validates that self-sensing signals in soft SMA-actuated soft robots can be used for proprioception and contact detection, and suggests a direction for integrating proprioception into soft robots without design compromises.
Future work could employ machine learning for enhanced accuracy.


\end{abstract}

\section{Introduction}
\label{Sec:introduction}
Robots made from soft materials are commonly claimed to demonstrate safer environmental interactions than their rigid counterparts \cite{Rus_Tolley_2015}. 
For intelligent control of these interactions, it is essential to estimate the robot's pose, forces, and contact states -- a process called \textit{proprioception} \cite{toshimitsu_sopra_2021}.
However, these robots' deformability complicates the integration of proprioceptive sensing \cite{mccandless_soft_2023}. 
Introducing dedicated force sensors can make the robot more fragile or lessen its natural compliance, which may defeat the purpose of softness \cite{alian2023soft}. 
Even when dedicated sensors are well-integrated or are soft themselves, they can be limited to contact detection at one location on the robot \cite{hellebrekers_soft_2019},
add design complexity with additional failure modes \cite{thuruthel2019soft}, or limit manipulation tasks with bulkiness \cite{mccandless2022cc,van_lewen_capacitive_2024}.
A robust, integrated solution is still lacking.

\begin{figure}[!t]
    \centering
    \includegraphics[width=\columnwidth]{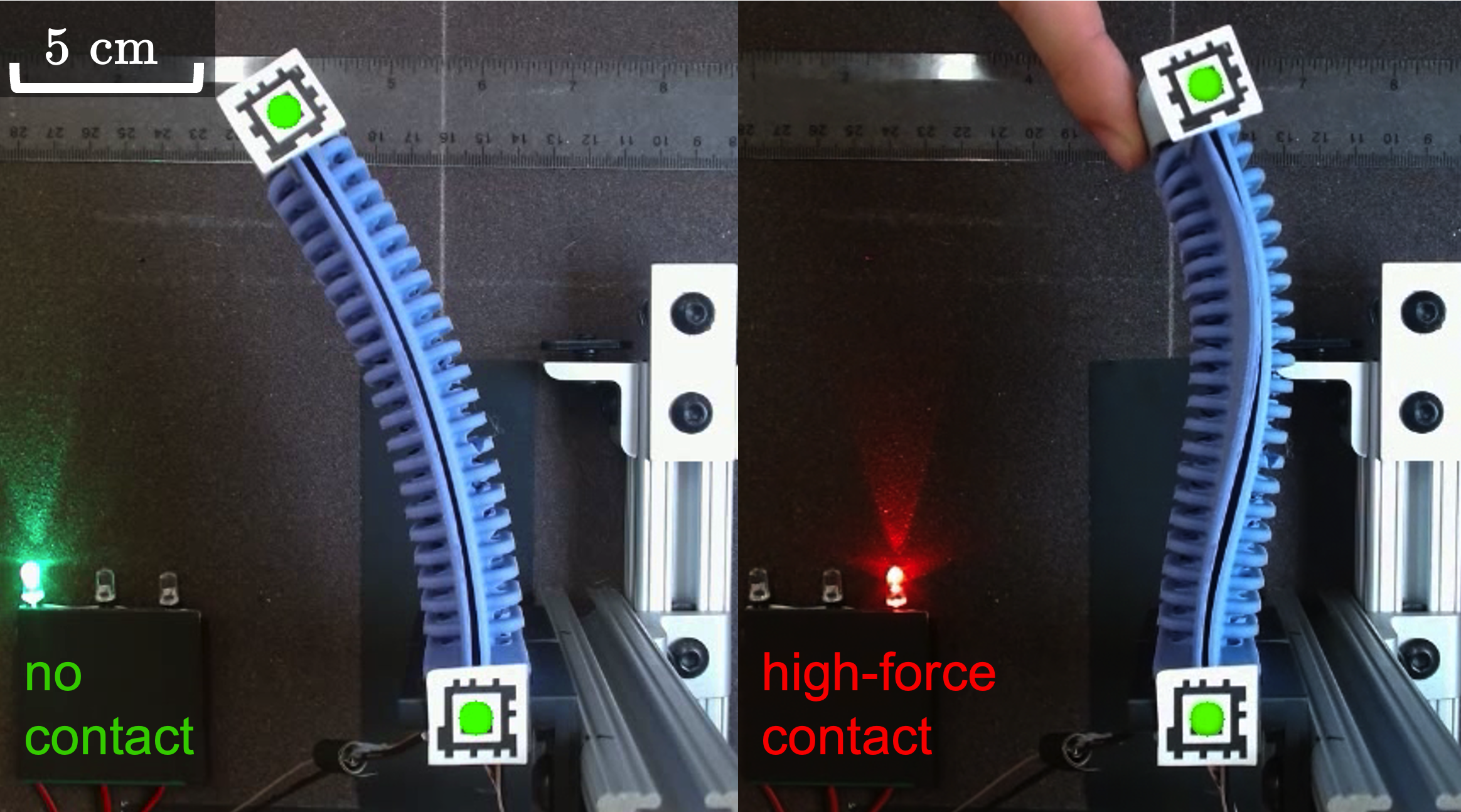}
    \caption{Our approach, using self-sensing of stress in a soft robot's artificial muscle actuators, can detect when external contact has occurred. No additional sensors are added to the soft limb's design.}
    \label{fig:first-page-shot}
    \vspace{-0.5cm}
\end{figure}

This work develops a proprioception approach for a soft robot limb using no additional sensors on the robot dedicated to that task.
We exploit a particular type of artificial muscle, the shape memory alloy (SMA) thermoelectric actuator, which can self-sense its stress through electrical resistance measurements \cite{prechtl2021model}.
Our insight is that this actuator stress maps to the soft robot's pose under no-contact conditions.
And if a true pose measurement is available, a difference between predicted vs. measured pose indicates that the robot has deformed under an external load (Fig. \ref{fig:first-page-shot}), thus detecting contact anywhere on the robot's body.

This paper contributes a proof-of-concept for this self-sensing of internal stress and external contact in an SMA-based soft robot limb.
Our scientific contributions demonstrate that electrical resistance of the thermoelectric SMA wire (when below its critical transition temperature) deconflicts the hysteresis in actuator temperature versus stress.
Additionally, we show that resistance is comparable to a temperature signal during contact detection.
Given a simple polynomial data fit as an estimator, we show pose errors of $15.3\%$ throughout the full operating range of the SMAs (up to $160^\circ$ C), and contact force prediction errors of $\approx 0.02$ N when below their critical transition temperature ($90^\circ$ C).

\begin{figure*}[t]
    \centering
    \includegraphics[width=\textwidth]{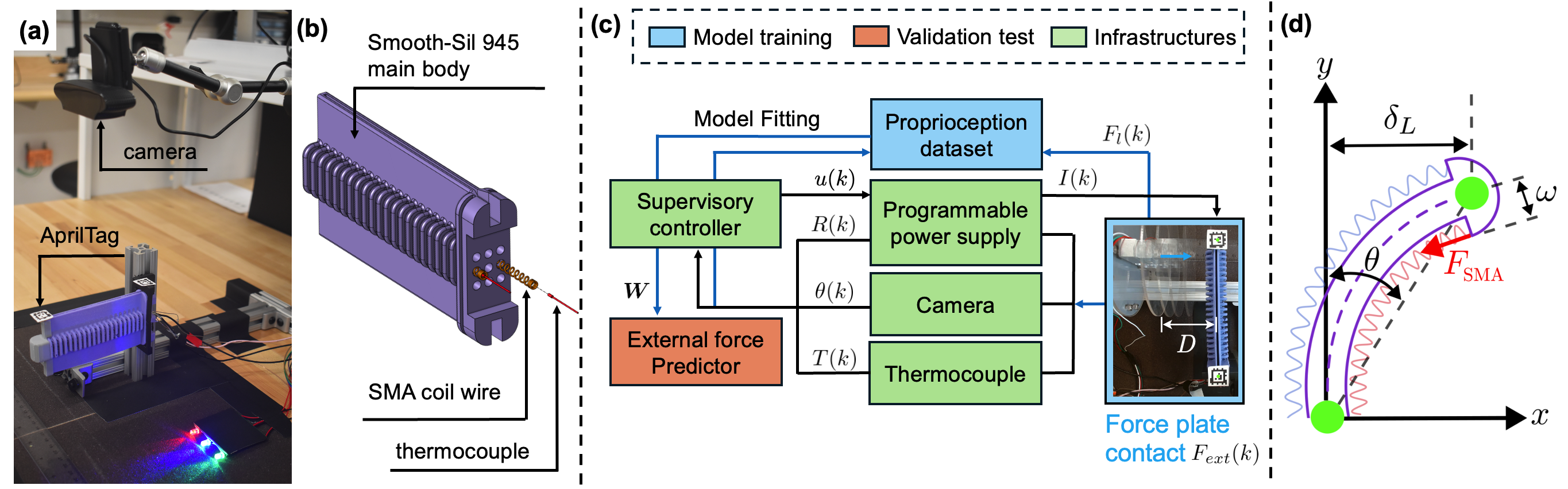}
    \caption{Architecture overview. Our existing limb design uses (a) computer vision for ground truth measurements of the robot's pose, where the limb includes (b) a silicone body actuated by shape memory alloy wire coils with a thermocouple affixed at the base. (c) Our data-driven approach includes our prior work in temperature-safe feedback control of these thermal SMA wires, with a data collection and sensing setup. (d) Our free-motion contact model uses a simple beam-bending approximation to map pose to actuator stress.}
    \label{fig:system_intro}
    \vspace{-0.3cm}
\end{figure*}

\subsection{Existing Approaches in Proprioception}


State-of-the-art in estimating a soft robot's pose and external loads commonly uses either dedicated sensors or bulky external hardware.
Dedicated sensors can be traditionally rigid and more mature, or softer and more exploratory, whether resistive \cite{tian2024multi}, capacitive \cite{kramer2011soft,atalay_batch_2017}, ionic \cite{annabestani2023bioinspired}, magnetic \cite{yue_active_2022}, or optical \cite{she2020exoskeleton,kim_optimizing_2014,faris2023proprioception}, among many others.
These have been used in combination with both finite-element models \cite{tian2024multi} and machine learning \cite{chin_machine_2020,truby2020distributed,soter_2018_proprio,ang_learning-based_2022,truby2020distributed} to estimate internal stresses or pose.
However, these approaches may require multiple sensors to infer properties of the whole robot \cite{kim_optimizing_2014} due to their localized readings, and may increase rigidity at each location. 
Increasing sensor density and surface area can address the issue \cite{lin20239dtact}, but at the cost of design tradeoffs.

External sensors can also be used to determine the pose of a soft robot and its contact forces, commonly via RGB-D cameras or motion trackers \cite{dellasantina2018dynamic} \cite{katzschmann2019dynamic}.
These high-quality measurements can similarly be used with either finite element models \cite{zhang2019calibration} or kinematic approximations \cite{della_santina_datadriven_2020} for pose and force estimation.
However, these techniques are not \textit{proprioception} in spirit, since proprioception implies sensors internal to the robot's body \cite{toshimitsu_sopra_2021}, and are therefore limited to predetermined and controlled environments.

\subsection{Actuator Self-Sensing and Proprioception}





Another method for proprioception is to use actuators that also serve as sensors, known as \textit{self sensing} \cite{gupta2019soft}.
Self-sensing requires fewer design constraints than dedicated sensors, even possibly no additional onboard hardware.
Dielectric elastomer actuators can self-sense via electrical changes versus deformation \cite{prechtl2024robot}, and liquid crystal elastomer actuators have a correlation between deformation and temperature \cite{wu2023modeling}, both allowing pose estimates of a soft robot.
However, both DEAs and LCEs face challenges in practical applications due to e.g. low force output, large driving voltages \cite{wang2022dielectric}, or slow speeds.
For pneumatic bellows, internal pressure sensors can be used to infer contact \cite{wang2023soft,toshimitsu_sopra_2021} but rely on bulky designs of pumps and valves.
\subsection{Shape Memory Alloy Actuator Self-Sensing and Soft Robot Integration}


Alternatively, shape memory alloy (SMA) artificial muscle actuators can be high force \cite{anderson2024maximizing} and high speed \cite{huang_highly_2019,Huang2018} with easy low-voltage integration onboard a soft robot \cite{patterson_untethered_2020}.
SMA actuators generate force by a reversible material-phase transformation when subjected to thermal stimuli. 
More importantly, SMAs can self-sense, as stress and strain relate to their temperature and material phase.
Prior work has demonstrated that under controlled lab conditions, SMA self-sensing can predict the actuator's displacement \cite{kennedy_2023_SMA_ML,Lee_2019_SMAposition,Prechtl_2021_SMAselfsense,ho_modeling_2013,cheng_modeling_2017} or the force it applies to a mechanism \cite{Pei_2023_ML_SMA_SS,Simone_2017_metalmusclegripper}, even enabling pose control via self-sensing \cite{ayvali_pulse_2014}.
However, these approaches struggle to translate to robots outside of precision test setups.

Our past work demonstrated that it is possible to integrate in-situ temperature measurements of SMAs in a soft robot, eliminating such test setups \cite{sabelhaus_-situ_2022,wertz_trajectory_2022}. 
We demonstrated pose predictions under no-load conditions \cite{sabelhaus_-situ_2022,wertz_trajectory_2022}, though with limited accuracy due to hysteresis \cite{pacheco2023comparison}.
However, it is known that electrical resistance of an SMA changes with its material phase \cite{lu_resistance_2020}, introducing an additional signal which could de-conflict hysteresis and predict stress.
To our knowledge, no prior work has integrated SMA stress self-sensing into a soft robot limb for proprioception and external contact detection.

\subsection{Approach and Outlook}

The remainder of this manuscript outlines our mathematical setup of our problem, hardware prototype, modeling approach, and testing.
We hypothesize that adding electrical resistance measurements of the SMA during operation will enable full proprioception.
We first validate pose estimation via actuator stress self-sensing, with no contact, before presenting contact detection as a separate question.
Our results are achieved using a simple least-squares polynomial fit to the self-sensing signals, prompting future work in machine learning techniques to improve contact prediction accuracy.

We utilize a computer vision system as a stand-in for in-situ pose measurements, which were previously obtained via a soft bending sensor in our existing designs \cite{wertz_trajectory_2022}, in order to reduce error and focus on our hypothesis of the SMA resistance self-sensing relationship.
However, since much of our past work \cite{patterson_robust_2022} and others \cite{koivikko2018screen}
contains such pose sensors as standard practice, we anticipate that our framework can be applied with only these already-onboard sensors.

\section{Soft Robot Limb Architecture}
\label{Sec:hardware and system design}


Since we hypothesize that self-sensing proprioception and contact detection is possible without design changes to a soft robot that uses SMA artificial muscles, this paper uses such a robot from our prior work \cite{wertz_trajectory_2022,anderson2024maximizing} (Fig.~\ref{fig:system_intro}(a)-(b)).
No additional on-board sensors are added.

\subsection{Actuator and Existing Sensors Design}


The existing prototype consists of 
a limb cast from silicone rubber (Fig.~\ref{fig:system_intro}(b)) that allows high force and rapid motion in a 2D plane. 
The limb contains a set of SMA wire coils (Dynalloy Flexinol 90$^\circ$C, 0.020” diam.) along each edge, which when heated via Joule heating through electric current, contract and cause bending.
We use a programmable power supply to apply a voltage per-actuator, taken as the control input to the robot ($\bu = V$), inducing corresponding currents $i$ which are recorded to a computer from the power supply.
In this paper, we isolate the limb to unidirectional motion with only a single actuator ($u \in \mathbb{R}^+$).
Using this commanded voltage and current reading from the power supply, we calculate a discrete-time series of resistance signals as the robot operates, $R(k) = V(k)/i(k)$.
We note that the resistance reading is therefore offboard and does not require a dedicated sensor in the robot.

Our prior manuscripts detail the computer vision system via AprilTags \cite{olson2011apriltag}, which calculates a bending angle $\theta(k)$, and a thermocouple at the base of the robot which reads an estimate $T(k)$ of the SMA's temperature \cite{sabelhaus_-situ_2022}.
Our most recent work \cite{anderson2024maximizing} has also integrated a force plate for external ground truth measurements of the limb's contact with an environment, $F_{ext}(k)$, so our datasets include $K$-many points $\{R(k), \theta(k), T(k), F_{ext}(k)\}^{K}$.




\subsection{Control System}\label{sec:controlsystem}
To calculate the input signals $u(k)$, we use a model-based temperature controller that monitors a nominal $u_{nom}(k)$, ensuring the robot does not overheat \cite{sabelhaus_safe_2022} during data collection i.e. ``motor babbling'' as $u_{nom}$ (Fig. \ref{fig:system_intro}(c)).
The temperature dynamics, extensively validated by our group, can be approximated as an affine scalar system, $T_{k+1} = a_1 T_k + a_2 u_k + a_3$. 
We calibrate this model using our approach from \cite{anderson2024maximizing}.

The supervisory controller ensures that temperatures remain below a maximum, $T(k) < T^{MAX}$.
To do so, we dynamically saturate the input $u_{nom}$ via  $u(k) = \min(u_{nom}(k), \gamma u^*(k))$, where $\gamma \in (0,1]$ is a discount factor for conservativeness. 
The dynamic saturation limit is $u^*(k) = \frac{1}{a_2}(T^{MAX}_{adj} - a_1 T(k) - a_3)$ and the discounted maximum temperature is $T^{MAX}_{adj} = (1/\gamma - a1((1-\gamma)/\gamma))T_{max} - a_3((1-\gamma)/\gamma)$, required as part of a proof of stability and invariance of the closed-loop system \cite{sabelhaus_safe_2022}.





\section{Actuator Force Self-Sensing}
\label{Sec:stress self-sensing}
Our first experiment tests the hypothesis that electrical resistance, together with \textit{in-situ} temperature measurements, can predict the robot's pose without hysteresis.

\subsection{Pose Proprioception Hypothesis}

Shape memory materials actuate via temperature change.
It is well-known \cite{Brinson1996,Patoor2006,hughes_preisach_1997} that the mechanics of induced stress in SMAs can be well-modeled as a lumped parameter system with four variables: $\sigma = f(\epsilon, T, \xi)$.
Here, $\sigma$ is stress in the material, $\epsilon$ is strain, $T$ is temperature, and $\xi \in [0,1]$ is the phase fraction of austenite vs. martensite \cite{cheng_modeling_2017}.
Although the material phase $\xi$ is difficult to measure, much prior work has demonstrated \cite{lan_accurate_2010,lu_resistance_2020,cui_modeling_2010,kim_sensorless_2012,bhargaw_performance_2021} that the electrical resistance $R$ of SMAs varies with $\xi$.
In addition, since our SMA coils have no appreciable cross-sectional area change, their applied force is $F_{SMA} \propto \sigma$.
Proprioception could occur, therefore, if a function $f(\cdot)$ exists so that $F_{SMA} = f(\epsilon, T, R)$.

\subsection{No-Contact Artificial Muscle Actuator Force Model}


We observe that in the case of no external contact on the limb, and under static equilibrium, 
the limb's bending angle $\theta(k)$ is uniquely determined by $F_{SMA}(k)$.
Furthermore, under the assumption of constant-curvature deformation in a soft robot \cite{webster2010design}, the limb's bending angle $\theta$ has a one-to-one geometric relationship with the strain $\epsilon$ of the SMA.





Our prior work has shown that a simple Euler-Bernoulli beam bending model sufficiently models this deformation for an SMA-powered soft limb\cite{patterson_robust_2022}.
Here, the SMA connection at the distal edge of the limb creates a moment at the tip.
The tip displacement at time $k$ of $\delta_{L}(k) =  L \sin^2(\theta(k))/\theta(k)$, determined by trigonometry of a constant-curvature arc (Fig.~\ref{fig:system_intro}(d)).
With a distance of $w$ between the SMA's attachment point and the limb's centerline, Euler-Bernoulli beam mechanics admit the analytical relationship:


\vspace{-0.3cm}
\begin{equation}\label{eqn:F_from_theta_beammoment}
    F_{SMA} = \left( \frac{4EI}{L^2w} \right) \delta_{L} = \left( \frac{4EI}{Lw} \right)  \left( \frac{\sin^2(\theta)}{\theta} \right),
\end{equation}

\noindent where $E$ and $I$ are the elastic modulus and area moment of inertia of the robot as a beam, and $L$ is limb's length.


Observe that eqn. (\ref{eqn:F_from_theta_beammoment}) is a function of only the limb bending angle $\theta$ under free motion. By defining the constant $\zeta = 4EI/(L^2w) \in \mathbb{R}$, eqn. (\ref{eqn:F_from_theta_beammoment}) for timestep $k$ can be written as $F_{SMA}(k) = \zeta \sin^2(\theta(k))/\theta(k)$.

For our prototype, the calibrated angle-to-stress coefficient is $\zeta=4.1767N/rad$, using $E=1.79N/(mm^2)$, $L=105$mm, and $w=3.5$mm, via geometry and material specifications for Smooth-Sil 945, and 
$I = bh^3/12$, where the limb's rectangular cross section is $b=60$ mm $h=3.5$ mm.
The model predicts $1.06N$ of the muscle force of the SMA when the bending angle is $15^{\circ}$, matching the experimental results in \cite{anderson2024maximizing} with a prediction error less than 10\%.



\subsection{Muscle Force and Limb Pose Estimation using SMA State}\label{sec:nocontact_model}


Eqn. (\ref{eqn:F_from_theta_beammoment}) suggests that predicting the pose $\theta(k)$ is equivalent to predicting the SMA force $F_{SMA}$ under no-contact, and that $\theta$ determines strain $\epsilon$.
We therefore seek a function $F_{SMA} = g(T,R)$, where $F_{SMA}$ is calculated from a measured $\theta$, simplifying the hypothesis.

Since the first-principles relationship between $\sigma$, $T$, and $R$ is complicated and highly nonlinear, we decide to take a data-driven approach to determine an estimator $\hat F_{SMA} = g(T,R)$. 
We collect a set of datapoints $d_f=\{\theta, T, R\}$ tuples during free movement of the limb, and convert to $\{F_{SMA}, T, R\}$ via eqn. (1).
To create these motions, we generate the motor babbling signal $u_{nom}$ from Sec. \ref{sec:controlsystem} as PI feedback on randomly selected bending angle setpoints, $\bar \theta$, as in \cite{sabelhaus_-situ_2022}, with a temperature limit $T_{MAX}=135^\circ$ C, justified by our recent work \cite{anderson2024maximizing}.
We chose ten setpoints between $10^\circ < \bar \theta < 40^\circ$ with multiple trials for a total dataset of $D_f=\{d_f(k) | k=1...600\}$ points.

Inspecting the collected data, we observe highly different $\{T,R\} \rightarrow F_{SMA}$ relationships at low temperatures versus high temperatures, diverging around $\approx 100^\circ$C (blue vs. orange in Fig.~\ref{fig:no_contact_surface_fit_3d}), which is around the critical transition temperature for our Nitinol alloy.
Visually, the relationship appears linear below that point, and quadratic afterwards.
We therefore manually split the dataset into $D_{cold} = \{d_f(k)|(T(k)<100^{\circ}C) \land (R(k)>1.7\Omega)\}$, and $D_{hot} = D_f \setminus D_{cold}$, and consider polynomial curve fits of various degrees for both sets alongside a switching strategy.


For a prediction using an $m$-degree polynomial at some test point $\bx = [T, R]^\top$, we construct an extended vector $\bX^{e}_m = \mathcal{L}(\bx, m)\in\mathbb{R}^{N_m\times1}$.
The operator $\mathcal{L}$ generates a vector of monomials of the form $[x^e_m]$, where $\{x^e_m = x_1^px_2^q...x_m^z,p,q,...,z | \text{sum}(p,q,...,z) = 0,...,m\}$.
For instance, with a quadratic fit ($m=2$), then $\bX^{e}_2 = [1, T, R, TR, T^2, R^2]^\top$.
Then, given a vector of weights for each monomial, $\bW\in\mathbf{R}^{N_m\times1}$, the polynomial prediction is $\hat F_{SMA} = \bW^\top \bX^{e}_m$.
To find the weights $\bW$ for a dataset $D_j = \{d_f(1)...d_f(J)\}$, we calculate $\bm{F_{SMA}}=[F_{SMA}(1),...,F_{SMA}(J)]^\top$ from eqn. (\ref{eqn:F_from_theta_beammoment}) and extended state matrix of all datapoints as $\bM^e_m = [\bX^{e}_m(1),...,\bX^{e}_m(J)]^\top \in \mathbb{R}^{J\times N_m}$.
The least-squares fit is then $\bW = (\bM^{e}_m)^\dag\bm{F_{SMA}}$, where $\dag$ is the pseudoinverse.
Applying this approach to $D_{hot}$ and $D_{cold}$ with $m=\{1,2\}$ with gives the weights $\bW_{hot}$ and $\bW_{cold}$.
Our predictor over the whole state space is then

\vspace{-0.3cm}
\begin{equation}
    \hat F_{SMA}(\bx) = \begin{cases}
        \bW_{cold}^\top \bX^e_m & \text{if} \; T<100^{\circ} \land R>1.7\Omega\\
        \bW_{hot}^\top \bX^e_m & \text{else}
    \end{cases} \label{eqn:Fhat_nocontact}
\end{equation}

\subsection{Experimental Result}

\begin{figure}[bt]
    \centering
    \includegraphics[width=1\columnwidth]{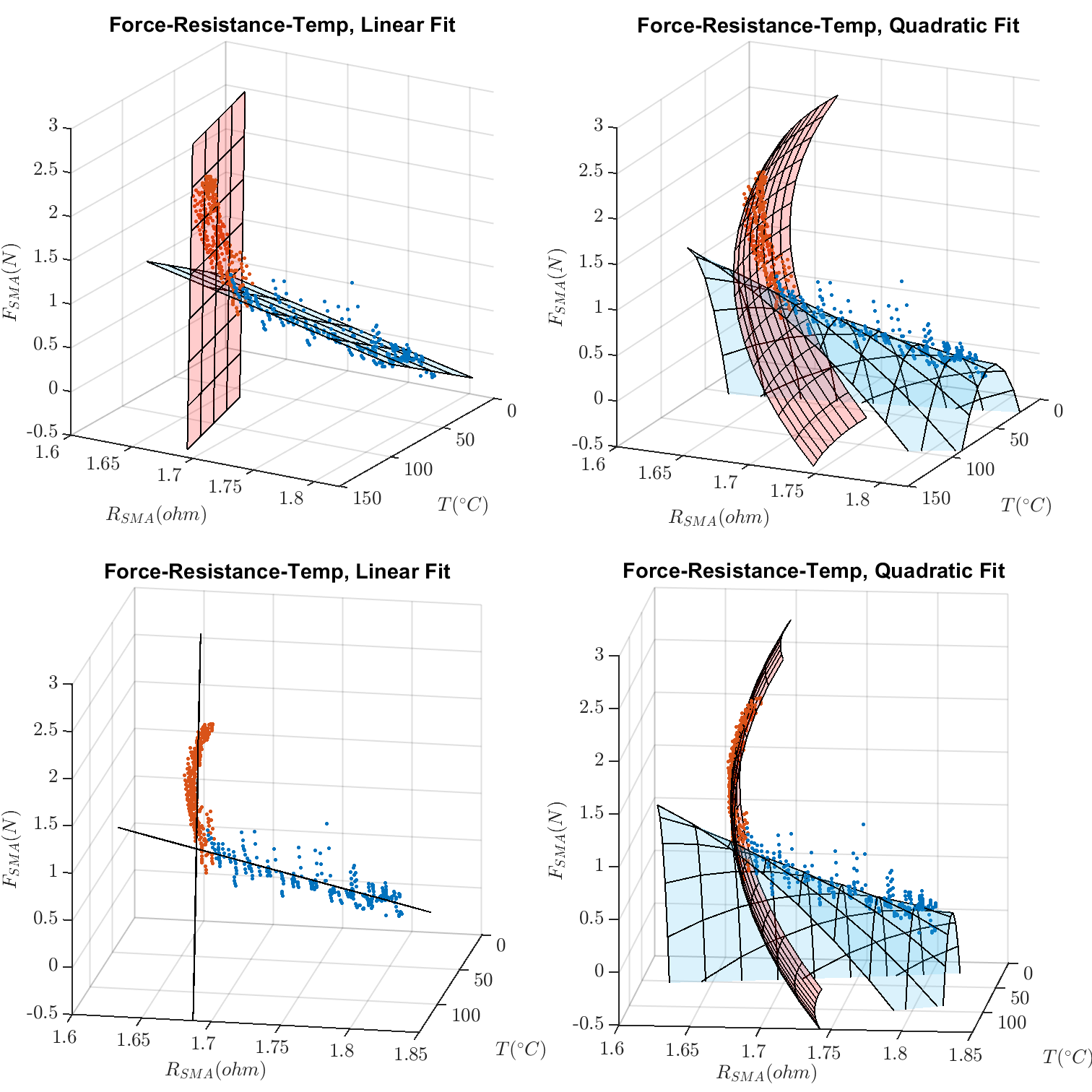}
    \caption{Both linear and quadratic regression models for no-contact actuator force self-sensing demonstrate that resistance deconflicts temperature hysteresis, given a split between hot (orange) and cold (blue) operating ranges per the SMA's critical transition temperature.}
    \label{fig:no_contact_surface_fit_3d}
    \vspace{-0.3cm}
\end{figure}

\begin{figure}[bt]
    \centering
    \includegraphics[width=\columnwidth]{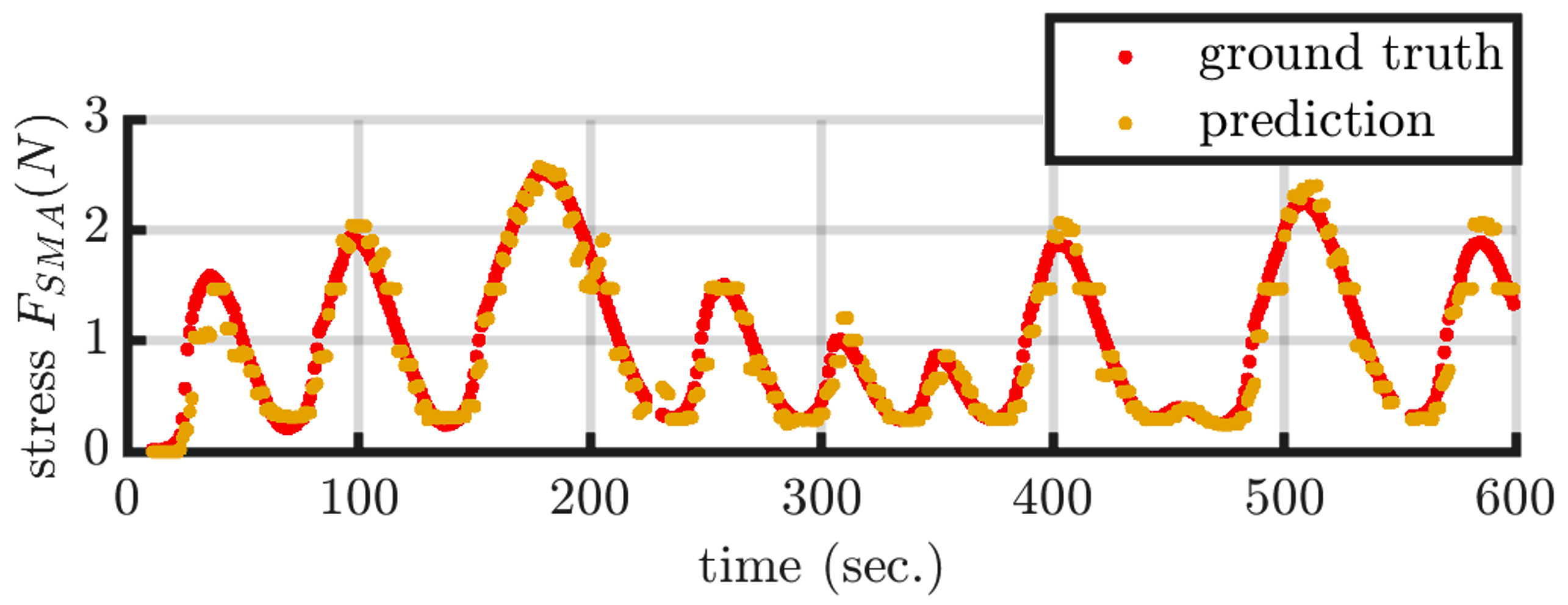}
    \caption{A snapshot of no-contact actuator force predictions shows a realistic trend using only self-sensing of resitance and in-situ temperature readings.}
    \label{fig:no_contact_stree_2d}
    \vspace{-0.5cm}
\end{figure}

Using a standard 3-fold cross validation, we obtained the $\bW_{hot}$ and $\bW_{cold}$ plotted as polynomial surfaces in Fig.~\ref{fig:no_contact_surface_fit_3d}, showing that a linear fit $(m=1)$ is acceptable for $D_{cold}$ but quadratic is needed for $D_{hot}$, and therefore $m=2$ is was used in the validation test.
A snapshot of time-series predictions over a test set in Fig.~\ref{fig:no_contact_stree_2d} visually confirms the quality of these actuator force (i.e., robot pose) predictions.
With the prediction error denoted $e(k) = \hat F_{SMA}(k)-F_{SMA}(k)$, the average error was $\bar e = \sum_{k}e(k)/(K-1)$ = $0.118 N$, which normalized as a percent was $\bar e_p = \sum_k e(k)/(F(k) (K-1)) = 15.28 \%$.
This is significantly improved over our prior work without resistance self-sensing \cite{pacheco2023comparison,sabelhaus_-situ_2022}.




\section{Contact Force Estimation}
\label{Sec:external force prediction}

For manipulation or locomotion tasks in soft robotics, estimation of external applied forces may be just as important as the robot's pose.
We have shown above that when the robot's pose change is only due to internal actuator stress, $F_{SMA}$, the mechanics can be accurately learned from self-sensing alone as $\hat F_{SMA} = g(T, R)$, where our proof-of-concept $g(\cdot)$ is eqn. (\ref{eqn:Fhat_nocontact}).
However, upon the addition of an external load $F_{ext}$ anywhere along the robot, the beam bending equations become statically indeterminate: the displacement $\delta_L(\theta)$ from eqn. (\ref{eqn:F_from_theta_beammoment}) becomes $\delta_L(\theta, F_{ext})$.
We therefore hypothesized that given an additional signal of the robot's pose (i.e., $\theta$ from another internal sensor), sufficient information exists to predict $F_{ext}$ during contact.




\subsection{Contact Force Prediction model}





In this setting, we now seek to predict the external force $F_{ext}$ from system state $\bx = [T, R, \theta]^\top$, in other words, a function $f(\cdot)$ such that $\hat F_{ext} = f(T, R, \theta)$. 
Motivated by the positive results during no-contact, we consider again polynomial fits on $\bx$.
Assume we want to derive a predictor using a dataset of motions that now contains intermittent environmental contact, $D_c=\{F_{ext}(k), T(k), R(k), \theta(k)\}, k=1\hdots K$.
As in Sec. \ref{sec:nocontact_model}, a $m$-degree polynomial prediction uses the extended state vector $\bX^{e}_m = \mathcal{L}(\bx, m)$ now with our $\bx = [T, R, \theta]^\top$.
The prediction is $\hat F_{ext}(\bx) = \bW^\top \bX^{e}_m$.
And as before, we can calculate the weights per least-squares, arranging the dataset as $\bm{F_{ext}}=[F_{ext}(1),...,F_{ext}(K)]^\top$ and $\bM^e_m = [\bX^{e}_m(1),...,\bX^{e}_m(K)]^\top$.
The fit is then $\bW = (\bM^{e}_m)^\dag\bm{F_{ext}}$.
For this setting, we extrapolate that a polynomial of $m=3$ is appropriate, since $\bx \in \mathbb{R}^3$ now.

\subsection{Contact Force Dataset and Model Fitting}

We generate a such data set $D_c$ using the external force plate setup from \cite{anderson2024maximizing}, per the procedure in Fig.~\ref{fig:system_intro}(c).
The force plate was positioned parallel to the soft limb at four fixed distances from the tip's neutral position, $\delta_L^{MAX} = \{ 20,30,40,50\}mm$.
We select four different SMA temperature limits $T^{MAX} = \{85,100,115,130\} ^\circ C$, motivated by the no-contact observation that resistance relationships change qualitatively around 90-100$^\circ$ C.
For each parameter combination, we manually design the $u_{nom}$ control signal's hold times at various $\bar \theta$ to span the state space of $\bx$ as broadly as possible.
We kept a small amount of current ($i > 0.2$A) at all times for our resistance calculation to be valid.
With 16 different parameter settings, we obtained over 400 minutes of active robot motion, for $K=24,000$ datapoints.



To test our hypothesis, we again use 3-fold cross validation on $D_c$.
However, we now split each training set into three separate models with different combinations of self-sensing states, to compare the benefits of including resistance measurements: $\{R,T,\theta\}$, $\{R,\theta\}$ and $\{T,\theta\}$.



\subsection{Experimental Results}

\begin{figure}[bt]
    \centering
    \includegraphics[width=\columnwidth]{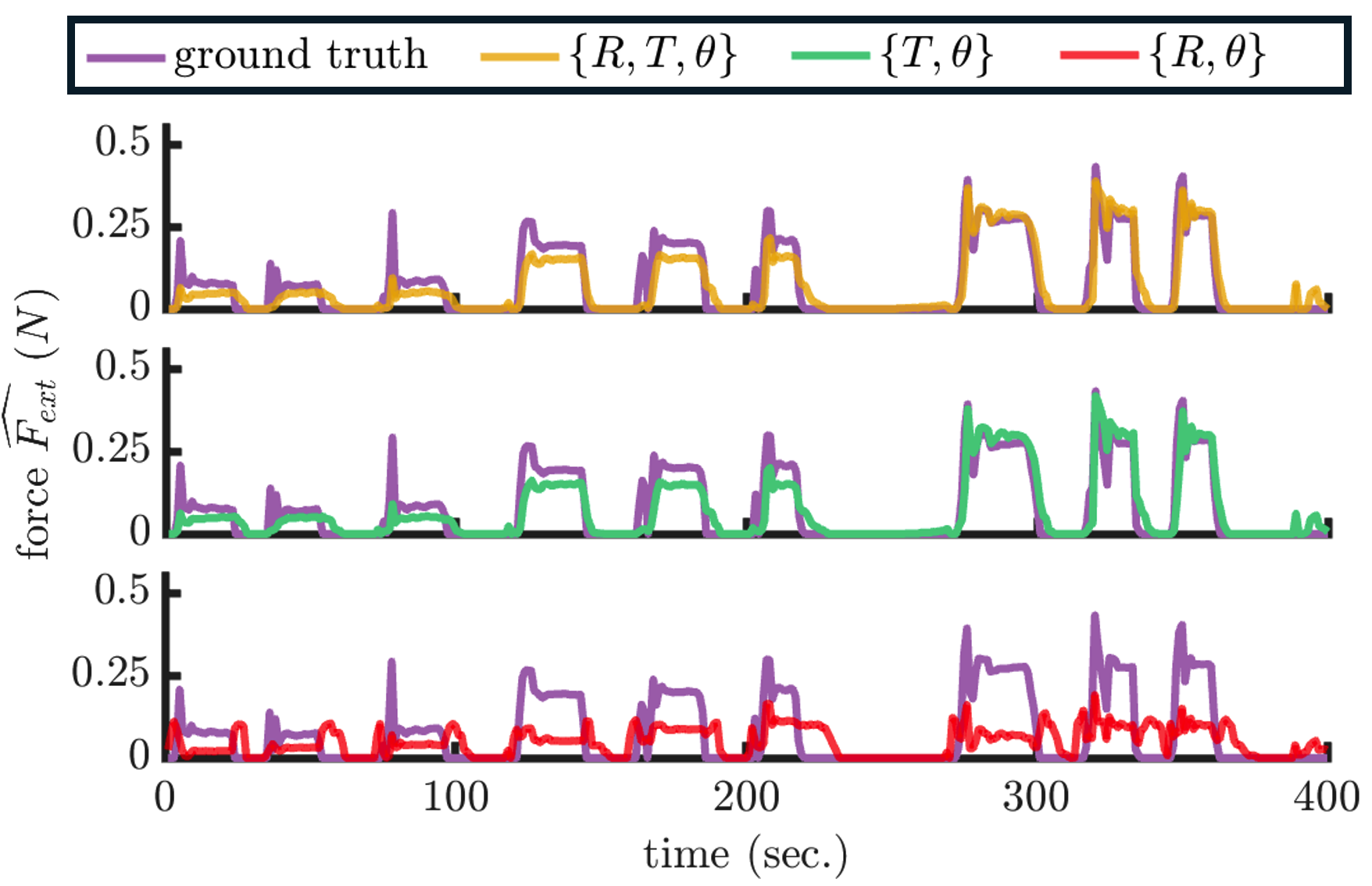}
    \caption{External force estimation with self-sensing signals shows that including electrical resistance $R$ in addition to temperature and pose ($T,\theta$) has minimal improvement. However, we notice that the $R,\theta$ model predicts nonzero forces at similar times as those with temperature, prompting the question: can $R$ substitute $T$ for contact detection?}
    \label{fig:3_params_combination_comp}
    \vspace{-0.3cm}
\end{figure}

Table \ref{tab:external_prediction_error} shows the average prediction error of the three external force predictors using our cubic polynomial fit ($m$=3) on the same testing set, and Fig.~\ref{fig:3_params_combination_comp} shows a snapshot of the models applied to the test set.


\begin{table}[ht]
    \centering    
    \caption{Avg. External Force Prediction Errors}
    \label{tab:external_prediction_error}
    \begin{tabular}{c | c c c c}
        Self-Sensing Signal Combination: & $\{R,T,\theta\}$ &  $\{R,\theta\}$ &  $\{T,\theta\}$   \\
        \hline
        Avg. Prediction Error ($N$):  & 0.0247 & 0.0250 & 0.0620 \\ 
    \end{tabular}
    \vspace{-0.3cm}
\end{table} 

The overall results demonstrate that external force estimation is indeed possible with our self-sensing signals.
Visually, both $\{R,T,\theta\}$ and $\{T,\theta\}$ have high-quality force proprioception.
Statistically, including resistance may lower the error during no-contact predictions, but visually in Fig.~\ref{fig:3_params_combination_comp}, there is little benefit (c.f. green versus orange time series).
Removing temperature signals appears to increase errors, yet all predictors seem to register nonzero values similarly to ground truth.
It is uncertain if this inconclusive result arises from our simplistic choice of estimator, or from the underlying physics of the problem.
Future work is needed to investigate the unsolved questions we mentioned above.


\section{Contact Detection and Operational Limits}
\label{Sec:binary contact detection}

\begin{figure*}[!t]
    \centering
    \includegraphics[width=\textwidth]{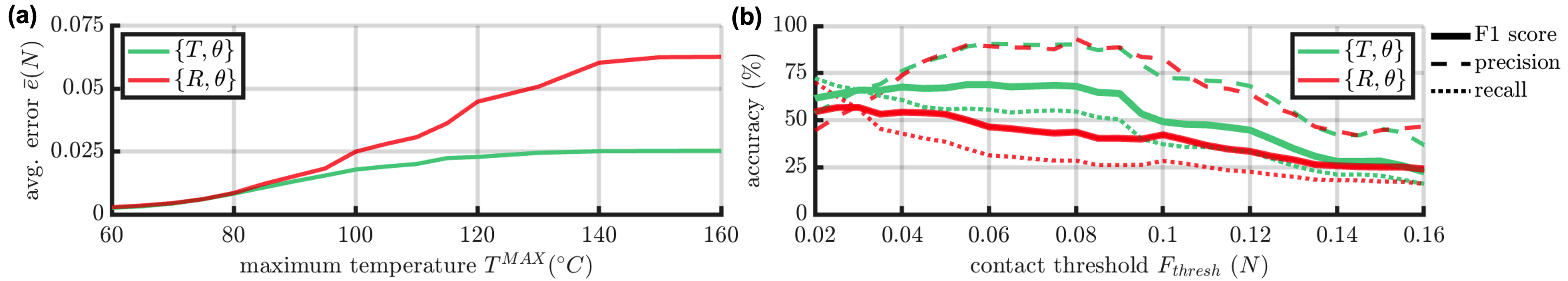}
    \caption{Contact force estimation errors are comparable between the self-sensing signals of temperature and pose $\{T,\theta\}$ versus electrical resistance and pose $\{R,\theta\}$ up to the $90^\circ$C critial transition temperature of the SMA (a). Under this operational limit, a binary contact detection classifier can be optimally calibrated with a simple threshold $F_{thresh}$ by taking the maximum of some statistic, e.g., F1 score or precision (b).}
    \label{fig:operational_limitation}
\end{figure*}


This manuscript's goal is validation of electrical resistance as a self-sensing signal for SMA-based soft robot proprioception, rather than an advanced predictor of $F_{ext}$.
However, we have observed nonzero contact forces in similar cases among all three predictors, so it may be possible that resistance is interchangeable with temperature in limited settings.
We therefore evaluate if a binary contact vs. no-contact state prediction is possible with self-sensing.
Along the way, we contribute a surprising discovery about our shape memory actuator: the limit where resistance is interchangeable with temperature also happens to be approximately the critical transition temperature (about $90^\circ$ C for our Nitinol).


\subsection{Binary Contact Detection and Calibration}

As a simple test, we introduce a classifier for binary detection of contact ($C=1$) or no contact $(C=0)$ of the soft limb against an environment, using a threshold on the estimated external force: $C = \{1 \; \text{if} \; \hat F_{ext} > F_{thresh}, \; \text{else} \; 0\}$.
We then seek to find an optimal $F_{thresh}$ for different $\hat F_{ext}$ models, and use the optimal classifiers to compare test statistics on $C$.
However, since electrical resistance appears to saturate around a certain temperature, this question is deeply intertwined with selecting a maximum temperature $T^{MAX}$ which is valid for each model.

We propose a data-driven procedure to iteratively determine $F_{thresh}$ and $T^{MAX}$, serving as a calibration method for the classifier, focusing on resistance as a substitute for temperature.
We run a series of regressions, selecting sub-datasets per temperature limit: $D_s(T^{MAX}) = \{d(k)|T(k)\in [0, T^{MAX}]\}$.
We pick different $T^{MAX}$ between $[60, 130] ^{\circ}C$ with a step size of 5$^\circ$ to build subsets, and cross-validate.

Fig.~\ref{fig:operational_limitation}(a) shows that electrical resistance can be used in place of temperature, $\{T,\theta\}$ versus $\{R,\theta\}$, for similar-quality $F_{ext}$ predictions up to a clear limit on temperature.
It was surprising that this $T^{MAX}$ coincided almost exactly with the no-contact scenario, and with the critical transition temperature: $90^\circ$ C.
We therefore focus on $\{R,\theta\}$ vs. $\{T,\theta\}$ models only below this operational limit of $90^\circ$ C.


Next, we select the optimal threshold $F^*_{thresh}$ by sweeping over a range of $F_{thresh}$ values, calculating classification statistics for each, and picking a pointwise maximum.
We use the standard statistical metrics of precision and recall to evaluate how well a contact detector matches the ground truth: \textit{Pre} = TP/(TP + FP) and \textit{Rec} = TP/(TP + FN), where TP is true positive, and FP (N) is false positive (negative).
Since there is a trade-off between precision and recall, we also calculate the F1-score, where F1=2\textit{Pre}*\textit{Rec}/(\textit{Pre}+\textit{Rec}).
Fig.~\ref{fig:operational_limitation}(b) shows how the F1 score, precision and recall evolves as the contact threshold increases (only considering $T^{MAX} = 90^\circ C$). 
From the curve in Fig.~\ref{fig:operational_limitation}, we set $F^*_{thresh} = \argmax F1(F_{thresh})$, i.e., the threshold with the largest F1 score. 
In our test, $\max(F1)$ occurs at $F_{thresh}^*=0.04$ N.

Table~\ref{tab:binary_prediction_accuracy} shows these three metrics for $F_{thresh}^*$, and as a comparison, the precision-optimal threshold at $F_{thresh}=0.08$ N.
In both cases and as is visible in Fig.~\ref{fig:operational_limitation}(b), the precision is similar between both $\{R,\theta\}$ and $\{T,\theta\}$, yet recall suffers slightly with $R$, causing a correspondingly lower F1 score.

\begin{table}[ht]
    \centering    
    \caption{Contact Classifier Prediction Statistics}
    \label{tab:binary_prediction_accuracy}
    \begin{tabular}{c | c | c c c}
        Signals & $F_{thresh}$ & Precision (\%) & Recall (\%) & F1 (\%) \\
        \hline
        $\{T,\theta\}$ & 0.04N & 76.24& 60.73 & 67.61 \\
        $\{R,\theta\}$ & 0.04N & 73.79 & 42.94 & 54.29 \\ 
        $\{T,\theta\}$ & 0.08N & 90.28 & 54.92 & 68.06 \\
        $\{R,\theta\}$ & 0.08N & 93.15 & 31.40 & 46.97 \\ 
    \end{tabular}
    \vspace{-0.5cm}
\end{table} 


\subsection{Real-time external contact detection}

\begin{figure}[t]
    \centering
    \includegraphics[width=\columnwidth]{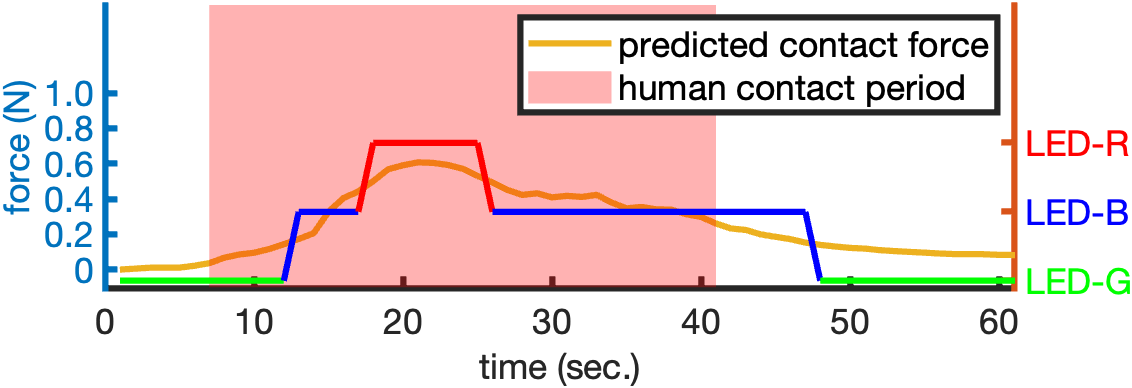}
    \caption{A demonstration for the human contact experiment and curves showing real-time model prediction and LED indication. }
    \label{fig:human_contact_result}
\end{figure}

Finally, we implemented the contact detector on hardware for a real-time detection demonstration. 
This time, instead of the rigid force plate in the data collection stage, a human user pushed on the robot with their finger, and an LED lights up depending on contact state (Fig.~\ref{fig:first-page-shot}).
This test used the $\{R,T,\theta\}$ estimate of $\hat F_{ext}$ as a demonstration.
In addition to the binary classifier for no contact (green) versus contact (blue), we added a threshold for high-force contact (red), calculated via our method as $0.1N$ and $0.5N$ respectively.
Fig.~\ref{fig:human_contact_result} provides a representative example of our human contact detector, demonstrating realistic behavior, though a delay occurs after the human removes their contact.



\section{Discussion, Conclusion, and Future Work}
\label{Sec:conclusion and discussion}

This paper demonstrates that electrical resistance self-sensing signals enable pose proprioception for an SMA-based soft robot limb, and complement existing signals when estimating external applied forces.
Using very simple polynomial regression models and a thresholding binary classifier, we show that it is possible to detect when a human contacts such a robot in real-time.
We propose a procedure to estimate the operational range of possible contact predictors, and a calibration scheme for the contact predictor.

This exploratory study validates our self-sensing concept, yet much work remains in practical implementation and accuracy improvements.
In particular, we have utilized ground-truth measurements (a computer vision system) for both training and as a signal during external force estimation.
Though it is standard practice to use CV in lab settings \cite{annabestani2023bioinspired}, work is underway to test these approaches with an existing prototype that includes an internal capacitive bending sensor that replaces the computer vision signal, creating true proprioception.
Additionally, our test conditions only included contact at one location in the robot.
Since our contact detection method did not make any assumptions about contact location, we plan to extend our tests to multi-contact settings.



Our approach demonstrates proof-of-concept force proprioception, but accuracy has room for improvement.
We believe that this misalignment could be improved by applying more advanced machine learning tools on our dataset, e.g., LSTMs \cite{sabelhaus_-situ_2022} or DNNs \cite{truby2020distributed}. 
These same methods may be necessary for multidirectional bending of the limb, or the use of multiple SMA muscles simultaneously.

Similarly, our results for estimating external forces $F_{ext}$ are inconclusive about the benefit of self-sensing electrical resistance in addition to temperature in-situ for SMA muscles.
It is unclear whether this is a limitation of our test setup and data collection (for example, if temperature was already too strongly correlated with contact), or with the choice of predictor (for example, if we were underfitting), or is a true physical phenomenon.
Future investigations will isolate this question by statistical analysis of the dataset, and learning models with larger numbers of parameters.

Finally, our future work will deploy an improved version of our proprioception method in human-robot collaboration tasks, where contact forces could be detected and limited to prevent injury to a human user.





\bibliographystyle{IEEEtran}
\bibliography{references}
\end{document}